\documentclass[11pt]{article}
\usepackage{tcolorbox}
\usepackage[preprint]{acl}
\usepackage{multirow}
\usepackage{microtype}
\usepackage{textcomp}
\usepackage{amsfonts}
\usepackage{times}
\usepackage{latexsym}
\usepackage{booktabs}
\usepackage{booktabs}
\usepackage{wrapfig}
\usepackage{wrapfig}
\usepackage{subcaption}
\usepackage[table]{xcolor}
\usepackage[T1]{fontenc}

\usepackage[utf8]{inputenc}
\usepackage{newunicodechar}
\newunicodechar{−}{\textminus}
\usepackage{amsmath}
\usepackage{microtype}

\usepackage{inconsolata}

\usepackage{graphicx}
\usepackage{xcolor}
\usepackage{listings}

\usepackage{algorithm}
\usepackage{algorithmic}
\usepackage{xcolor}
\definecolor{lightpurple}{HTML}{F4F0FF} 
\definecolor{sftblue}{HTML}{4B6FA5}  
\definecolor{rlred}{HTML}{C0504D}

\newcommand{\ours}{SimpleOCR}
\title{\ours: Rendering Visualized Questions to Teach MLLMs to Read}

\author{Yibo Peng$^{1,2\dagger}$\thanks{Equal Contribution. $^{\dagger}$Work was done during the internship at UNC}, Peng Xia$^{1*}$, Ding Zhong$^{1,3\dagger*}$, Kaide Zeng$^{1*}$, Siwei Han$^{1}$ \\ \textbf{Yiyang Zhou$^{1}$, Jiaqi Liu$^{1}$, Ruiyi Zhang$^{4}$, Huaxiu Yao$^1$}\\ $^1$UNC-Chapel Hill, $^2$Carnegie Mellon University, $^3$University of Michigan, $^4$Adobe Research \\ \small \texttt{yibop@andrew.cmu.edu,dingdd@umich.edu,\{pxia,kdzeng,huaxiu\}@cs.unc.edu}}

\begin{document}
\maketitle
\begin{abstract}
Despite the rapid advancements in Multimodal Large Language Models (MLLMs), a critical question regarding their visual grounding mechanism remains unanswered: do these models genuinely ``read'' text embedded in images, or do they merely rely on parametric shortcuts in the text prompt? In this work, we diagnose this issue by introducing the Visualized-Question (VQ) setting, where text queries are rendered directly onto images to structurally mandate visual engagement. Our diagnostic experiments on Qwen2.5-VL reveal a startling capability-utilization gap: despite possessing strong OCR capabilities, models suffer a performance degradation of up to 12.7\% in the VQ setting, exposing a deep-seated ``modality laziness.'' To bridge this gap, we propose \ours, a plug-and-play training strategy that imposes a structural constraint on the learning process. By transforming training samples into the VQ format with randomized styles, \ours\ effectively invalidates text-based shortcuts, compelling the model to activate and optimize its visual text extraction pathways. Empirically, \ours\ yields robust gains without architectural modifications. On four representative OOD benchmarks, it surpasses the base model by 5.4\% and GRPO based on original images by 2.7\%, while exhibiting extreme data efficiency, achieving superior performance with 30x fewer samples (8.5K) than recent RL-based methods. Furthermore, its plug-and-play nature allows seamless integration with advanced RL strategies like NoisyRollout to yield complementary improvements. Code is available at \href{https://github.com/aiming-lab/SimpleOCR}{\texttt{https://github.com/aiming-lab/SimpleOCR}}.
\end{abstract}

\section{Introduction}
\label{sec:intro}
Multimodal Large Language Models (MLLMs) have achieved remarkable progress in visual reasoning by integrating vision encoders with large language models \cite{liu2023visual,bai2023qwen,bai2025qwen2,hurst2024gpt,comanici2025gemini}. 
Central to this capability is optical character recognition (OCR), i.e., the ability to extract and interpret text embedded in images, which underpins performance on chart understanding \cite{masry2022chartqa,wang2024charxiv}, document analysis \cite{mathew2021docvqa,han2025mdocagent,mathew2022infographicvqa}, and geometry-centric reasoning \cite{lu2021inter,lu2023mathvista}.
While current MLLMs achieve strong performance on standalone OCR benchmarks, a fundamental question remains underexplored: \emph{do these models actually leverage their OCR capabilities when solving downstream tasks?}

To investigate this, we introduce a controlled diagnostic intervention called the \emph{visualized-question} (VQ) format.
In standard evaluation, models receive questions via text, which may allow reasoning based on linguistic priors or parametric shortcuts rather than visual evidence.
In the VQ setting, we render the question text directly onto the image and provide only a generic instruction (e.g., ``\texttt{Please answer the question in the image}''), forcing the model to ground its reasoning in visual text.
If a model fully utilizes its OCR capabilities, performance under both settings should be comparable.
However, our experiments reveal a striking \emph{capability--utilization gap}.
\begin{figure}[t]
    \centering
    \subfloat[Visualized-Question (VQ) Format]
    {
        \includegraphics[width=0.9\linewidth]{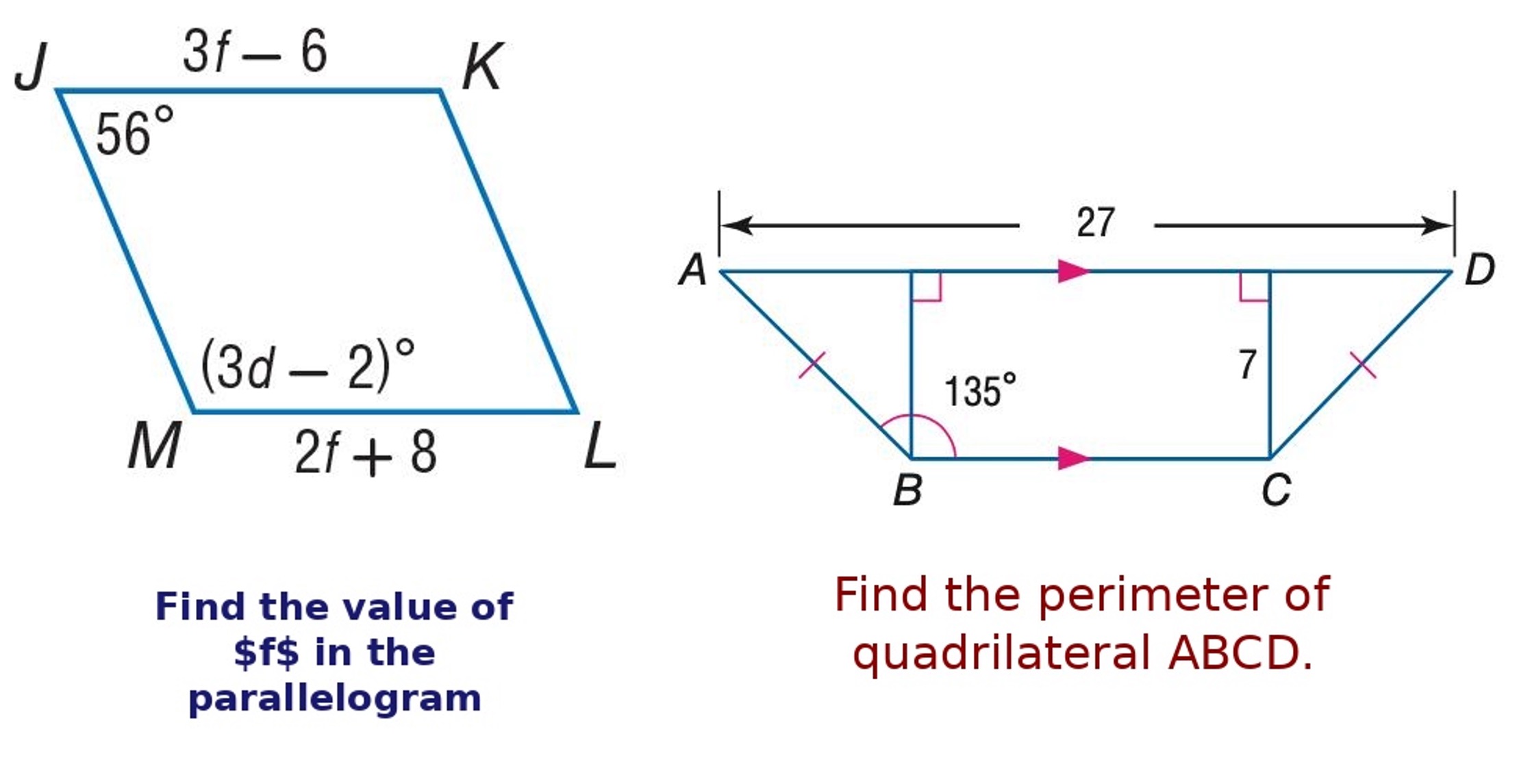}
        \label{fig:case_format}
    }
    \vspace{-0.5em}
    \par\medskip 
    \subfloat[Capability-Utilization Gap on Benchmarks]{
        \includegraphics[width=0.9\linewidth]{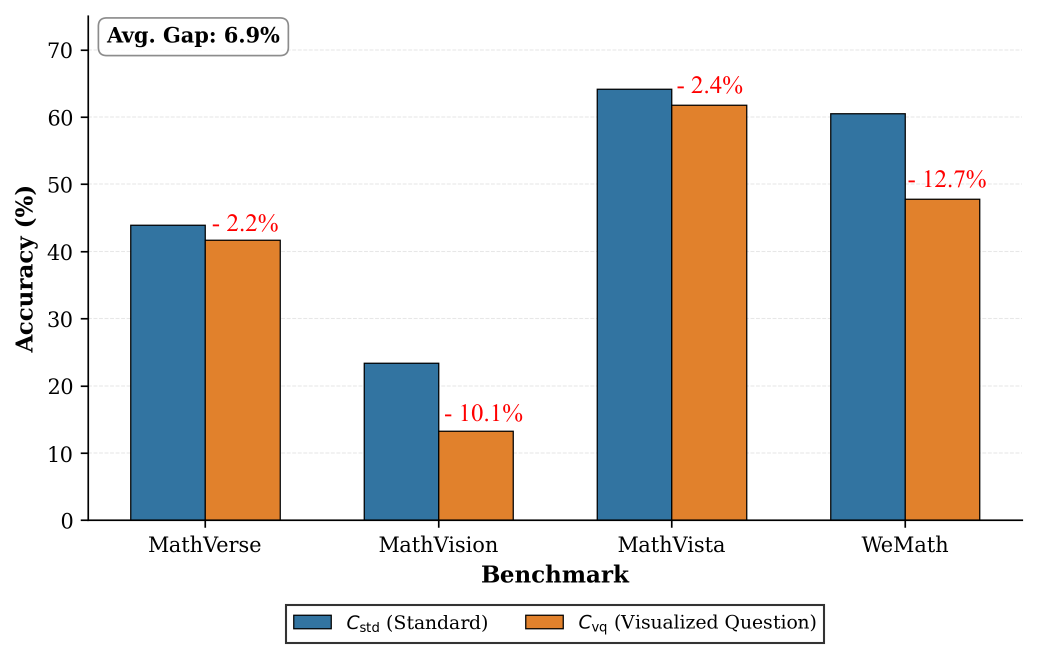}
        \label{fig:gap_chart}
    }
    \caption{(a) \textbf{Visualized-Question (VQ) Format.} We render the question text into the image as the only question source, removing text-channel shortcuts and requiring visual reading. (b) \textbf{Capability--Utilization Gap.} On Qwen2.5-VL-7B, performance drops markedly under VQ versus standard inputs, indicating that OCR capability is not reliably utilized during reasoning.}

    \label{fig:motivation}
\end{figure}

As shown in Figure~\ref{fig:motivation}, Qwen2.5-VL-7B suffers substantial degradation under the VQ setting, with an average absolute drop of 6.9\% across four multimodal reasoning benchmarks and a maximum drop of 12.7\% on WeMath~\cite{qiao2024wemath}.
This phenomenon aligns with recent observations of ``modality laziness'' \cite{lin2023revisiting,fu2025hidden,yao2025rethinking}, where models systematically underweight visual evidence when informative text prompts are available.

Motivated by this diagnosis, we propose \textbf{\ours}, a training strategy that addresses this gap through \emph{structural constraint}. Rather than auxiliary losses or architectural modifications \cite{yu2025perception,cao2025ground,sarch2025grounded}, \ours\ operates purely through input transformation: all training samples are converted to VQ format with randomized visual styles, eliminating text-based shortcuts entirely.
Notably, \ours\ introduces zero additional computational overhead or inference latency. By embedding questions directly into the visual space, it forces the model to decode image-based prompts prior to reasoning, thereby drastically improving OCR-based understanding. As a plug-and-play strategy, \ours\ can be seamlessly incorporated into any VLM training framework, enhancing model robustness and reasoning by enriching the visual distribution of training data.

Empirically, \ours\ induces robust performance gains across both in-domain (ID) and out-of-distribution (OOD) scenarios. When trained on Geo3K and MMK12, \ours\ achieves a 6.6\% improvement over the base model on ID test sets, and achieves 8.5\% compared to GRPO based on original images. The generalization capability of our approach is substantiated by results on challenging OOD benchmarks. On MathVerse, MathVision, MathVista, WeMath, and HallusionBench, \ours\ surpasses the base model by 5.4\% and GRPO based on original images by 2.7\%. Notably, \ours\ exhibits extreme data efficiency: with only 8.5K training samples, it outperforms RL-based methods \cite{zhang2025r1,yang2025r1} that require over 260K samples, demonstrating a 30x reduction in data dependency. Furthermore, \ours\ is a plug-and-play strategy that requires no modifications to the model architecture or training paradigms. It integrates seamlessly with existing VLM training frameworks. For instance, when combined with RL methods like NoisyRollout \cite{liu2025noisyrollout}, it yields complementary gains, confirming that \ours\ enhances a unique and orthogonal dimension of multi-modal reasoning.

Our primary contribution is \ours, a plug-and-play training strategy designed to bridge the OCR \textit{capability–utilization} gap. By imposing structural constraints, \ours\ forces models to actively engage with visual text, effectively addressing the performance degradation (up to 12.7\%) seen when text shortcuts are removed. Empirical results across multiple multimodal reasoning benchmarks demonstrate that our approach significantly enhances out-of-distribution generalization. Furthermore, we verify the effectiveness of our structural components and demonstrate the broad compatibility of \ours\ with existing multimodal architectures.

\section{Related Work}

\paragraph{Reinforcement Learning for MLLMs.}
Reinforcement Learning from Verifiable Rewards (RLVR) advances multimodal reasoning by utilizing programmatic signals rather than subjective preferences, extending the RLHF paradigm~\cite{ouyang2022training,yu2024rlhf,wang2025pixel,tu2025position,xia2025mmedagent,xia2025agent0,liu2025agent0vl,su2025thinking,xia2026skillrl,yang2025reliable}.
The GRPO algorithm~\cite{shao2024deepseekmath} has powered frontier models like DeepSeek-R1~\cite{guo2025deepseek}, with recent adaptations refining the framework through diverse mechanisms.
Specifically, R1-Onevision~\cite{yang2025r1} and Vision-R1~\cite{huang2025vision} optimize cross-modal formalization and training dynamics, respectively, while R1-VL~\cite{zhang2025r1} and VLAA-Thinker~\cite{chen2025sft} introduce step-wise rewards and mixed perception-cognition signals.
To enhance stability, MM-Eureka~\cite{meng2025mm} and ThinkLite-VL~\cite{wang2025sota} employ data-centric strategies such as rejection sampling and MCTS-based selection.
Then NoisyRollout~\cite{liu2025noisyrollout} targets policy diversity by mixing distorted trajectories.
However, these methods primarily focus on logical derivation or robustness, lacking explicit constraints to enforce visual text reading against shortcut learning.
\paragraph{Visual Grounding in Text-Rich Contexts.}
The paradigm for text-rich understanding has shifted from modular OCR pipelines to unified end-to-end architectures~\cite{bai2025qwen2,zeng2025glm,li2024llava,zhang2025skywork}.
To circumvent resolution constraints, Monkey~\cite{li2024monkey} and TextMonkey~\cite{liu2024textmonkey} introduced patch-division strategies, while VisInContext~\cite{wang2024leveraging} leveraged visual tokens to efficiently scale context length.
Subsequently, architectures like GOT~\cite{wei2024general} and Donut~\cite{blecher2023nougat} unified perception and reasoning.
Current state-of-the-art models, including Qwen2.5-VL~\cite{bai2025qwen2}, MiniCPM-V 4.5~\cite{yu2025minicpm}, and HunyuanOCR~\cite{team2025hunyuanocr}, leverage large-scale OCR corpora~\cite{geng2025webwatcher} and native-resolution ViTs~\cite{dosovitskiy2020image} to handle complex layouts.
Despite these advances in \textit{capability acquisition}, a critical dichotomy remains: models possess strong OCR capabilities but suffer from systematic ``modality laziness''~\cite{fu2025hidden,yao2025rethinking}, failing to utilize visual evidence during reasoning.
Unlike prior works, our work targets \textit{capability utilization}, ensuring the model actively grounds its reasoning in visual text evidence.

\begin{figure*}[t]
    \centering
    \includegraphics[width=\linewidth]{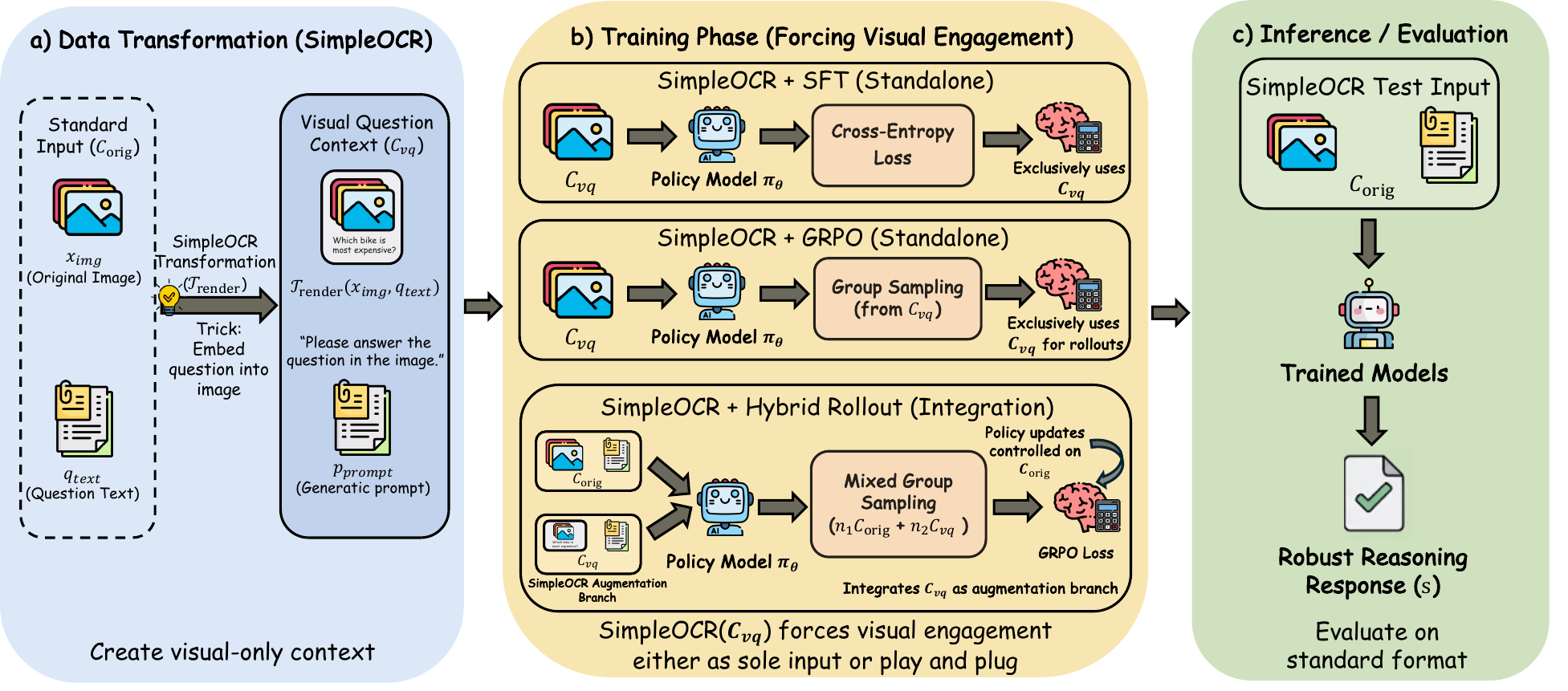}
    \caption{The SimpleOCR framework. During training, all inputs are transformed into visual question contexts $C_{vq}$, where question text is rendered onto images. This structurally eliminates text-based shortcuts and forces visual OCR engagement. At inference, models trained this way demonstrate robust performance on standard inputs $C_{orig}$. The method integrates seamlessly as an augmentation branch in existing RL frameworks.}
    \label{fig:framework}
\end{figure*}
\section{Preliminaries}
In this section, we will provide a brief overview of MLLMs and GRPO algorithm. 
We build upon Group Relative Policy Optimization (GRPO)~\cite{shao2024deepseekmath}, a reinforcement learning framework designed to improve the reasoning ability of large language models. 

Given a multimodal question $q$, consisting of an image $x_{\text{img}}$ and a text prompt $q_{\text{text}}$, the policy model $\pi_\theta$ generates a reasoning response $o$. For each question $q$, GRPO samples a group of $G$ candidate responses $\{o_1, o_2, \ldots, o_G\}$ from the old policy $\pi_{\theta_{old}}$. Each response $o_i$ is assigned a reward $r_i$ (e.g., from a reward model or rule-based verifier). The group-relative advantage $\hat{A}_i$ for each response is then computed by:
\begin{equation}
\hat{A}_i = \frac{r_i - \frac{1}{G} \sum_{j=1}^G r_j}{\text{std}(r_1, \ldots, r_G)},
\label{eq:group_adv}
\end{equation}
which centers and normalizes the rewards within the group, effectively removing question-level biases.

The policy model $\pi_\theta$ is updated by maximizing the GRPO objective, which incorporates a PPO-style clipped surrogate loss and a KL divergence penalty against a frozen reference model $\pi_{\text{ref}}$:
\begin{equation}
\begin{aligned}
\mathcal{L}_{\text{GRPO}}(\theta) &= \mathbb{E}_{q \sim \mathcal{D}, \{o_i\} \sim \pi_{\theta_{\text{old}}}} \Bigg[ \frac{1}{G} \sum_{i=1}^G \bigg( \\
& \min \left( r_i(\theta) \hat{A}_i, \text{clip}\left(r_i(\theta), 1-\epsilon, 1+\epsilon\right) \hat{A}_i \right) \\
& - \beta D_{KL}(\pi_\theta \| \pi_{\text{ref}}) \bigg) \Bigg],
\end{aligned}
\label{eq:grpo_obj}
\end{equation}
where $r_i(\theta) = \frac{\pi_\theta(o_i|q)}{\pi_{\theta_{\text{old}}}(o_i|q)}$ denotes the probability ratio. The hyperparameters $\epsilon$ and $\beta$ represent the clipping range and the KL divergence penalty weight, respectively. By bypassing the value function and utilizing group-relative advantages, GRPO significantly optimizes memory usage and training efficiency while maintaining robust performance.
\section{\ours: Addressing the Gap Through Visual Question Training}

\subsection{Visual Question Setting}
Given a training sample $S = (\mathbf{x}_{\text{img}}, q_{\text{text}})$, we define two informationally equivalent yet structurally distinct input contexts.

\paragraph{Standard Context $C_{\text{orig}}$.} This context preserves the conventional multimodal schema, $C_{\text{orig}} = (\mathbf{x}_{\text{img}}, q_{\text{text}})$, where the question is provided via the text channel.

\paragraph{Visual Question Context $C_{\text{vq}}$.} To structurally enforce visual grounding, we introduce a transformation $\mathcal{T}_{\text{render}}$ that embeds the semantic content of $q_{\text{text}}$ directly into the visual modality:
\begin{equation} 
    C_{\text{vq}} = \left( \mathcal{T}_{\text{render}}(\mathbf{x}_{\text{img}}, q_{\text{text}}), \quad p_{\text{prompt}} \right)
\end{equation}
where $p_{\text{prompt}}$ is a generic instruction (e.g., ``\texttt{Answer the question in the image}''). By removing $q_{\text{text}}$ from the text channel, $C_{\text{vq}}$ eliminates the possibility of text-based shortcuts, making visual text reading structurally necessary.

As detailed in Algorithm~\ref{alg:render}, $\mathcal{T}_{\text{render}}$ appends the question text to a canvas region below the original image, ensuring all original visual features are preserved. To prevent the model from overfitting to specific layouts, we employ a randomized rendering strategy: parameters such as font family (with CJK support), color, and size (dynamically scaled between 18--42pt) are sampled stochastically during training. This diversity ensures that the learned OCR capabilities are robust to varying visual presentations.

\begin{algorithm}[t]
\caption{Visual Question Rendering ($\mathcal{T}_{\text{render}}$)}
\label{alg:render}
\definecolor{codegreen}{rgb}{0.0,0.5,0.4}
\begin{algorithmic}[1]
\small
\STATE \textcolor{codegreen}{\# x: original image, q: question text}
\STATE \textbf{def} render(x, q):
\STATE \hspace{1em} \textcolor{codegreen}{\# Sample random style (language-aware)}
\STATE \hspace{1em} font, color $\gets$ random\_style()
\STATE \hspace{1em} size $\gets$ random.randint(18, 42)
\STATE
\STATE \hspace{1em} \textcolor{codegreen}{\# Wrap text and create canvas}
\STATE \hspace{1em} lines $\gets$ wrap(q, width=x.width, size=size)
\STATE \hspace{1em} h $\gets$ len(lines) $\times$ line\_height(size)
\STATE \hspace{1em} canvas $\gets$ Image.new((x.width, x.height + h), white)
\STATE
\STATE \hspace{1em} \textcolor{codegreen}{\# Paste original image and draw text}
\STATE \hspace{1em} canvas.paste(x, (0, 0)) 
\STATE \hspace{1em} draw(canvas, lines, font, size, color, y=x.height)
\STATE \hspace{1em} \textbf{return} canvas
\end{algorithmic}
\end{algorithm}

\begin{algorithm}[t]
\footnotesize
\caption{\ours\ Training Strategy}
\label{alg:simpleocr_grpo}
\begin{algorithmic}[1]
\REQUIRE Dataset $\mathcal{D}$, Policy $\pi_{\theta}$, Reference $\pi_{\theta_0}$, Renderer $T_{\text{render}}$
\ENSURE Optimized Policy $\pi_{\theta}$

\FOR{each batch $(x_{\text{img}}, q_{\text{text}}, a) \in \mathcal{D}$}
    \STATE \textcolor{blue}{\textit{$\triangleright$ 1. Construct Visual Question Context}}
    \STATE $x_{\text{render}} \leftarrow T_{\text{render}}(x_{\text{img}}, q_{\text{text}})$
    \STATE $C_{\text{vq}} \leftarrow (x_{\text{render}}, p_{\text{prompt}})$

    \STATE \textcolor{blue}{\textit{$\triangleright$ 2. Group Sampling (Visual Exploration)}}
    \STATE Sample $G$ outputs from visual context: $\{s_1, \ldots, s_G\} \sim \pi_{\theta}(\cdot | C_{\text{vq}})$
    
    \STATE \textcolor{blue}{\textit{$\triangleright$ 3. Advantage Computation}}
    \FOR{$k=1$ to $G$}
        \STATE Compute reward $r_k$ comparing $s_k$ to ground-truth $a$
    \ENDFOR
    \STATE $\hat{A}_k = \frac{r_k - \text{mean}(\mathbf{r})}{\text{orig}(\mathbf{r}) + \epsilon}$ \quad \textcolor{gray}{// Group-relative advantage}
    
    \STATE \textcolor{blue}{\textit{$\triangleright$ 4. Policy Update}}
    \STATE Compute GRPO loss on $C_{\text{vq}}$: $\mathcal{L} = -\frac{1}{G}\sum_{k=1}^{G} \left[ \hat{A}_k \log \pi_{\theta}(s_k | C_{\text{vq}}) - \beta \mathbb{D}_{\text{KL}} \right]$
    \STATE Update $\theta$ using gradient descent
\ENDFOR
\end{algorithmic}
\end{algorithm}

\subsection{Training Strategy}

\ours\ trains models exclusively on visual question format. All training samples undergo the $\mathcal{T}_{\text{render}}$ transformation which is no mixing of standard and visual question formats during training. This design eliminates text channel shortcuts entirely, forcing every training update to engage the visual text reading pathway.

Our approach is implemented purely as data preprocessing via $\mathcal{T}_{\text{render}}$, requiring no architectural changes and no modification to standard training objectives. For RL training, as illustrated in Alg.~\ref{alg:simpleocr_grpo}, we follow the standard GRPO algorithm while conditioning generation on $C_{\text{vq}}$: we first construct $x_{\text{render}}$ and $C_{\text{vq}}$, sample a group of $G$ responses, compute rewards and group-relative advantages, and update the policy using the GRPO objective with the KL regularizer unchanged.

Critically, while training uses exclusively $C_{\text{vq}}$, evaluation employs standard format $C_{\text{orig}}$. This forces models to develop format-agnostic reasoning capabilities rather than format-specific patterns, learning to extract and process question content regardless of presentation modality.

\subsection{Plug-and-Play Integration}

Beyond standalone training, SimpleOCR integrates seamlessly into existing training frameworks. We demonstrate this with NoisyRollout \cite{liu2025noisyrollout}.

NoisyRollout employs a hybrid rollout strategy: for each sample, it generates $n_1$ rollouts from clean images $(\mathbf{x}_{\text{img}}, q_{\text{text}})$ and $n_2$ rollouts from perturbed images $(T_\alpha(\mathbf{x}_{\text{img}}), q_{\text{text}})$, where $T_\alpha$ applies image distortion with strength $\alpha$. All rollouts contribute to computing group-relative advantages, improving policy exploration and visual robustness.

We integrate SimpleOCR by substituting the perturbation branch with visual question samples. Specifically, we generate $n_1$ rollouts from the standard context $C_{\text{orig}}$ and $n_2$ rollouts from the visual question context $C_{\text{vq}}$. All rollouts contribute to group-relative advantage computation as in standard NoisyRollout. Policy updates remain conditioned on $C_{\text{orig}}$ following NoisyRollout's original design. This integration requires no algorithmic modifications, as we simply substitute one augmentation strategy for another. The combination proves effective because the two methods target orthogonal objectives: NoisyRollout enhances visual robustness through image perturbations, while SimpleOCR specifically addresses OCR utilization through visual text reading.
\section{Experiments}

\subsection{Experiment Settings}

\paragraph{Dataset.}
We train on Geometry3K~\cite{lu2021inter} (2.1K instances) and MMK12~\cite{meng2025mm} (6.4K instances), totaling 8.5K instances. 

\paragraph{Evaluation.}
We evaluate on two dimensions: (1) \emph{in-domain} performance on Geometry3K and MMK12 test sets, and (2) \emph{out-of-distribution} generalization on MathVerse~\cite{zhang2024mathverse}, MathVision~\cite{wang2024measuring}, MathVista~\cite{lu2023mathvista}, and HallusionBench~\cite{guan2024hallusionbench}.
We additionally evaluate on OCR-intensive benchmarks: InfographicVQA~\cite{mathew2022infographicvqa} (InfoVQA) and ChartQA~\cite{masry2022chartqa}.
All evaluations utilize greedy decoding, followed by a hybrid judging pipeline combining symbolic verification (Math-Verify\footnote{https://github.com/huggingface/Math-Verify}) and LLM-based assessment (GPT-4o~\cite{hurst2024gpt}). We detail the full protocol in Appendix~\ref{app:eval_protocol}.

\subsection{Main Results}

\begin{table*}[ht]
    \centering
    \setlength{\tabcolsep}{4pt}
    \resizebox{\textwidth}{!}{
    \begin{tabular}{lcccccccccc} 
    \toprule
    & & \multicolumn{3}{c}{\textbf{In-Domain}} & \multicolumn{5}{c}{\textbf{Out-of-Distribution}} \\ 
    \cmidrule(lr){3-5} \cmidrule(lr){6-10}
    {Method} & {Data Size} & {Geo3K} & {MMK12} & {Avg.} & {MathVerse} & {MathVision} & {MathVista} & {HallusionBench} & {Avg.} \\
    \midrule
    \multicolumn{10}{c}{\textit{Open-source Baselines (SFT / General)}} \\
    \midrule
    InternVL-2.5-8B-Instruct* \cite{chen2024expanding} & - & - & - & - & 39.5 & 19.7 & 64.4 & 67.3 & 47.7 \\ 
    LLaVA-OneVision-7B* \cite{li2024llava}& - & - & - & - & 26.2 & - & 63.2 & 48.4 & - \\
    Kimi-VL-16B* \cite{team2025kimi} & - & - & - & - & 44.9 & 21.4 & \textbf{68.7} & 66.2 & 50.3 \\ 
    Mulberry-7B* \cite{yao2024mulberry} & - & - & - & - & - & - & 63.1 & - & - \\
    Math-LLaVA~\cite{shi2024math} & \textcolor{sftblue}{360K} & - &- & -& 22.9 & 15.7 & 46.6 & - & - \\
    \midrule
    \multicolumn{10}{c}{\textit{RL-Optimized Models (R1-series)}} \\
    \midrule
    R1-VL-7B \cite{zhang2025r1} & \textcolor{sftblue}{260K}+\textcolor{rlred}{10K} & 34.3 & 39.1 & 36.7 & 37.5 & 19.1 & 61.6 & 62.8 & 45.3 \\ 
    R1-OneVision-7B \cite{yang2025r1} & \textcolor{sftblue}{155K}+\textcolor{rlred}{10K} & 37.4 & 47.2 & 42.3 & 43.6 & 20.9 & 63.1 & 65.6 & 48.3 \\ 
    ThinkLite-7B-VL \cite{wang2025sota} & \textcolor{rlred}{1.1K} & 38.8 & 56.8 & 47.8 & 46.7 & 24.2 & 66.9 & 66.1 & 51.0 \\ 
    VLAA-Thinker-7B \cite{chen2025sft}& \textcolor{rlred}{25K} & 37.6 & 56.7 & 47.2 & \underline{46.9} & \underline{24.4} & \underline{67.6} & 68.1 & 51.8 \\ 
    MM-Eureka-8B*~\cite{meng2025mm} & \textcolor{rlred}{15K}  & -& -&- & 40.4 & 22.2 & 67.1 & 65.3 & 48.8 \\ 
    \midrule
    \multicolumn{10}{c}{\textit{Our Methods}} \\
    \midrule
    Qwen2.5-VL-7B-Instruct \cite{bai2025qwen2} & - & 37.6 & 53.6 & 45.6 & 43.9 & 23.4 & 64.2 & 68.2 & 49.9 \\ 
    \quad + GRPO & \textcolor{rlred}{\textbf{8.5K}} & \textbf{44.3} & \underline{61.9} & \textbf{53.1} & 46.4 & 22.5 & 66.9 & \underline{68.9} & \underline{51.2} \\ 
    \rowcolor{lightpurple}
    \quad + SimpleOCR & \textcolor{rlred}{\textbf{8.5K}} & \underline{43.4} & \textbf{62.3} & \underline{52.9} & \textbf{47.7} & \textbf{24.9} & \textbf{68.7} & \textbf{69.1} & \textbf{52.6} \\ 
    \bottomrule
    \end{tabular}}
\caption{Performance on mathematical reasoning and visual perception benchmarks. Models marked with ``*'' are cited from original papers. \textbf{Bold} and \underline{underlined} numbers indicate the best and second-best performance, respectively. Data sizes for SFT and RL are respectively marked in \textcolor{sftblue}{blue} and \textcolor{rlred}{red}. }
    \label{tab:main_results}
\end{table*}

\paragraph{Robust Transfer via Zero-Shot Generalization.}
SimpleOCR trains exclusively on VQ inputs but evaluates on standard inputs, creating a severe distributional shift that rigorously tests visual capability.
Rather than suffering the expected degradation from format mismatch, SimpleOCR achieves robust zero-shot transfer.
As shown in Table~\ref{tab:main_results}, it matches the baseline's in-domain performance (52.9\% vs. 53.1\%) while strictly outperforming it on out-of-distribution generalization (52.6\% vs. 51.2\%).
This transfer is most potent on visually demanding tasks like MathVision, where we observe a 10.7\% gain.
These results prove that the model has not merely memorized the VQ format, but has internalized a fundamental visual text extraction capability that persists even when text shortcuts are restored.

\paragraph{Gains Correlate with Visual-Text Dependency.} The performance improvements are structurally non-uniform. MathVision exhibits the most significant boost (24.9\% vs. 22.5\%), followed by MathVista (68.7\% vs. 66.9\%) and MathVerse (47.7\% vs. 46.4\%). Crucially, these benchmarks share a dependency on visual information density: they require extracting critical data or text embedded directly within figures.
In contrast, performance slightly regresses on Geometry3K (43.4\% vs. 44.3\%), a benchmark governed more by abstract geometric logic than by visual text reading. This divergence confirms that SimpleOCR specifically sharpens the visual-text extraction pathway rather than offering a generic reasoning boost. We consider this a strategic trade-off: a marginal dip in pure geometry is exchanged for robust generalization on tasks where visual grounding is paramount.

\begin{figure}[t]
    \centering
    \includegraphics[width=\linewidth]{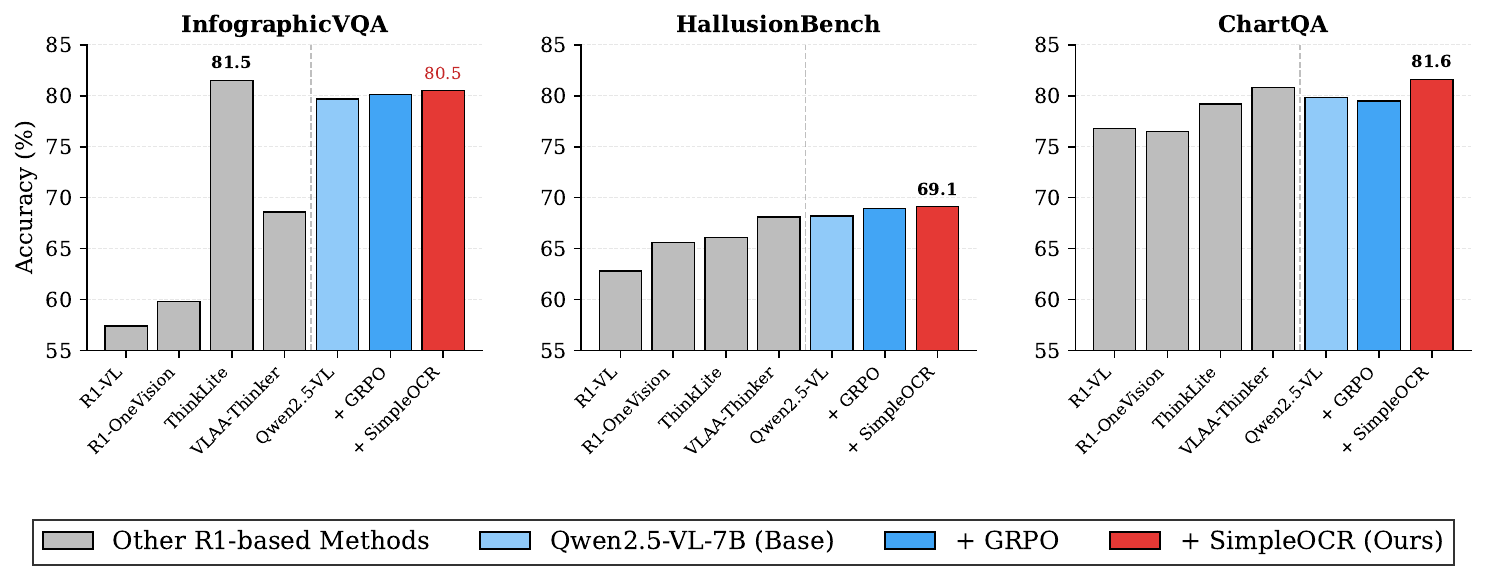}
    \caption{Performance on OCR-intensive benchmarks. SimpleOCR demonstrates superior performance, achieving 81.6\% on ChartQA and 69.1\% on HallusionBench.}
    \label{fig:ocr_benchmark}
    \vspace{-1em}
\end{figure}

\paragraph{Superiority on OCR-Intensive Benchmarks.}
Figure~\ref{fig:ocr_benchmark} (see Appendix Table~\ref{tab:ocr_results}) confirms that SimpleOCR excels on tasks requiring explicit visual text recognition.
On ChartQA, while standard GRPO slightly degrades performance (79.8\% $\rightarrow$ 79.5\%), SimpleOCR reverses this trend, reaching 81.6\%.
Consistent improvements are observed on InfographicVQA and HallusionBench, reaching 80.5\% and 69.1\%, respectively.
This establishes a clear hierarchy of efficacy: gains are pronounced on OCR-centric tasks (e.g., ChartQA) and visually grounded math (e.g., MathVision), but negligible on pure geometry (e.g., Geometry3K).
This distribution confirms that SimpleOCR functions as a targeted enhancer of visual-text utilization rather than a generic regularizer.

\subsection{Analysis}

\begin{table*}[t]
    \centering
    \caption{\textbf{Analysis of Integration \& Scalability:} SimpleOCR integrates seamlessly with HybridRollout across model scales. The combination yields consistent gains, particularly on the 3B model, validating that SimpleOCR (focused on text reading) and HybridRollout (focused on visual robustness) are orthogonal and complementary.}
    \label{tab:analysis_noisyrollout}
    \resizebox{\textwidth}{!}{%
    \begin{tabular}{lccccccccc} 
    \toprule
    & \multicolumn{3}{c}{\textbf{In-Domain}} & \multicolumn{5}{c}{\textbf{Out-of-Distribution}} \\ 
    \cmidrule(lr){2-4} \cmidrule(lr){5-9}
    \textbf{Method Configuration} & {Geo3K} & {MMK12} & {Avg.} & {MathVerse} & {MathVision} & {MathVista} & {Hallusion} & {Avg.} \\
    \midrule    
    Qwen2.5-VL-3B-Instruct & 26.0 & 45.9 & 36.0 & 32.5 & 18.2 & 50.8 & 59.1 & 40.2 \\ 
    \quad + GRPO (baseline) ($n=6$) & \underline{35.1} & 51.2 & 43.2 & 36.6 & 18.5 & 53.0 & 58.6 & 41.7 \\ 
    \quad + SimpleOCR ($n=6$) & \textbf{36.1} & \textbf{53.6} & \textbf{44.9} & \underline{40.4} & \underline{20.1} & \underline{53.3} & \textbf{61.7} & \underline{43.9} \\ 
    \quad + HybridRollout ($n_1=3$, $n_2=3$) & 34.8 & \textbf{53.6} & \underline{44.2} & \textbf{41.4} & \textbf{20.6} & \textbf{57.3} & \underline{58.7} & \textbf{44.5} \\ 
    \midrule
    Qwen2.5-VL-7B-Instruct & 37.6 & 53.6 & 45.6 & 43.9 & 23.4 & 64.2 & 68.2 & 49.9 \\ 
    \quad + GRPO (baseline) ($n=6$) & \textbf{44.3} & 61.9 & \textbf{53.1} & 46.4 & 22.5 & 66.9 & \underline{68.9} & 51.2 \\ 
    \quad + SimpleOCR ($n=6$) & \underline{43.4} & \underline{62.3} & \underline{52.9} & \textbf{47.7} & \textbf{24.9} & \textbf{68.7} & \textbf{69.1} & \textbf{52.6} \\ 
    \quad + HybridRollout ($n_1=3$, $n_2=3$) & 41.1 & \textbf{65.0} & \textbf{53.1} & \underline{47.6} & \textbf{24.9} & \textbf{68.7} & 68.0 & 52.3 \\ 
    \bottomrule
    \end{tabular}%
    }
\end{table*}

\paragraph{Plug-and-Play Compatibility.}
Table~\ref{tab:analysis_noisyrollout} demonstrates the compatibility of SimpleOCR with advanced training strategies like NoisyRollout~\cite{liu2025noisyrollout}. 
On Qwen2.5-VL-7B, SimpleOCR outperforms the GRPO baseline by 2.7\%. This trend is consistent at the 3B scale: SimpleOCR delivers a 5.3\% boost in average OOD accuracy, which is further amplified by the inclusion of NoisyRollout.
The consistent gains confirm that the methods target distinct reasoning dimensions: SimpleOCR provides semantic grounding, while NoisyRollout improves perceptual robustness. This orthogonality validates SimpleOCR as a flexible plug-and-play augmentation compatible with existing training paradigms.

\paragraph{Consistency Across Model Scales.}
We further investigate scaling behavior in Table~\ref{tab:analysis_noisyrollout}. On Qwen2.5-VL-7B, SimpleOCR delivers a robust 2.7\% over the GRPO baseline (52.6\% vs. 51.2\%), validating its efficacy beyond small-scale models.
While the gain margin naturally narrows compared to the 3B model (an expected consequence of \emph{performance saturation} in larger models), the consistent positive trajectory confirms that ``modality laziness'' is a fundamental architectural tendency irrespective of capacity.
SimpleOCR effectively mitigates this tendency regardless of scale, serving as a scalable corrective mechanism.

\subsection{Ablation Study}
\begin{figure}[t]
    \centering
    \includegraphics[width=0.9\columnwidth]{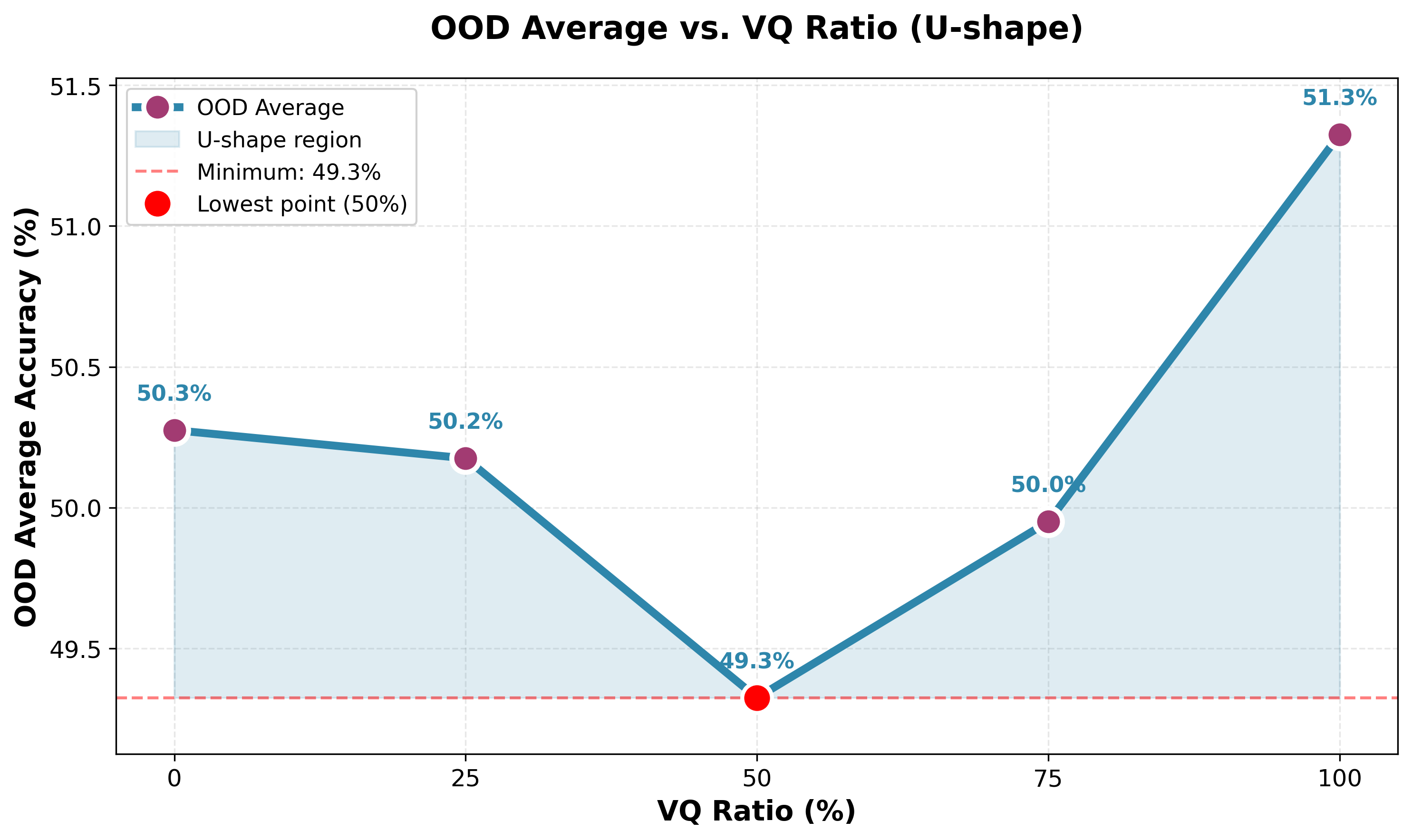} 
    \caption{The ``U-Shaped'' Optimization Conflict. We report the average performance across four representative OOD benchmarks. The mixed strategy (50\% VQ) results in a net performance loss, illustrating that contradictory modality signals hinder generalization.}
    \label{fig:ood_avg_u_shape}
\end{figure}

\paragraph{Optimization Conflict in Mixed Strategies.}
To better understand the interaction between standard inputs and VQ training, we evaluated a mixed strategy (Partial Exposure). Figure~\ref{fig:ood_avg_u_shape} reveals a distinct U-shaped performance trajectory.
On average across four representative OOD benchmarks (detailed in Appendix Table~\ref{tab:ablation_full_appendix}), the mixed setting (50\% VQ) unexpectedly dips below the baseline (49.3\% vs. 50.3\%), creating a generalization valley. This degradation is particularly pronounced on reasoning-heavy tasks like WeMath ($-4.4\%$) and MathVista ($-2.8\%$).

We attribute this to a fundamental optimization conflict. When exposed to mixed formats, the model receives contradictory learning signals: standard inputs encourage reliance on the text encoder (the path of least resistance), while VQ inputs demand active visual engagement. Rather than converging on a robust joint strategy, the model oscillates between these modalities, failing to master either.
The \ours\ (100\% VQ) setting resolves this by enforcing a structural constraint. By completely blocking text-based shortcuts, the model is compelled to optimize the visual extraction pathway. Paradoxically, this ``forced commitment'' yields representations that are modality-agnostic, enabling superior zero-shot transfer (51.3\% average accuracy).

\begin{table}[t]
    \centering
    \caption{Ablation on rendering style. Randomization prevents overfitting to specific visual patterns. (Note: ``Random style'' corresponds to the full \ours\ method used in main results.)}
    \label{tab:ablation_style}
    \resizebox{\columnwidth}{!}{%
    \begin{tabular}{lcccccc}
    \toprule
    & \multicolumn{2}{c}{\textbf{In-Domain}} & \multicolumn{4}{c}{\textbf{Out-of-Distribution}} \\
    \cmidrule(lr){2-3} \cmidrule(lr){4-7}
    \textbf{Rendering Strategy} & {Geo3K} & {MMK12} & {M-Verse} & {M-Vision} & {M-Vista} & {WeMath} \\
    \midrule
    Fixed style & 41.4 & 61.3 & 46.9 & 23.4 & 65.9 & 61.6 \\
    Random style & \textbf{43.4} & \textbf{62.3} & \textbf{47.7} & \textbf{24.9} & \textbf{68.7} & \textbf{64.0} \\
    \bottomrule
    \end{tabular}%
    }
\end{table}

\paragraph{Robustness via Randomization.}
Table~\ref{tab:ablation_style} validates the efficacy of our randomized rendering strategy. Compared to a static rendering style (e.g., fixed font and color), applying stochastic styles (varying font, size, and color) yields consistent gains, most notably a $2.8\%$ improvement on MathVista and $2.4\%$ on WeMath.
We attribute the limitations of the fixed setting to \emph{feature overfitting}. When text always appears with a deterministic visual style, the model tends to memorize low-level texture cues (e.g., specific font patterns) rather than performing generalizable OCR.
Randomization disrupts these shortcuts. By diversifying the stylistic presentation, we compel the model to actively decode text regardless of its visual variations. This strategy effectively prevents the model from relying on nuisance variables (such as font type or color), ensuring that the learned grounding capability is genuinely robust.

\begin{figure*}[t]
    \centering
    \includegraphics[width=\linewidth]{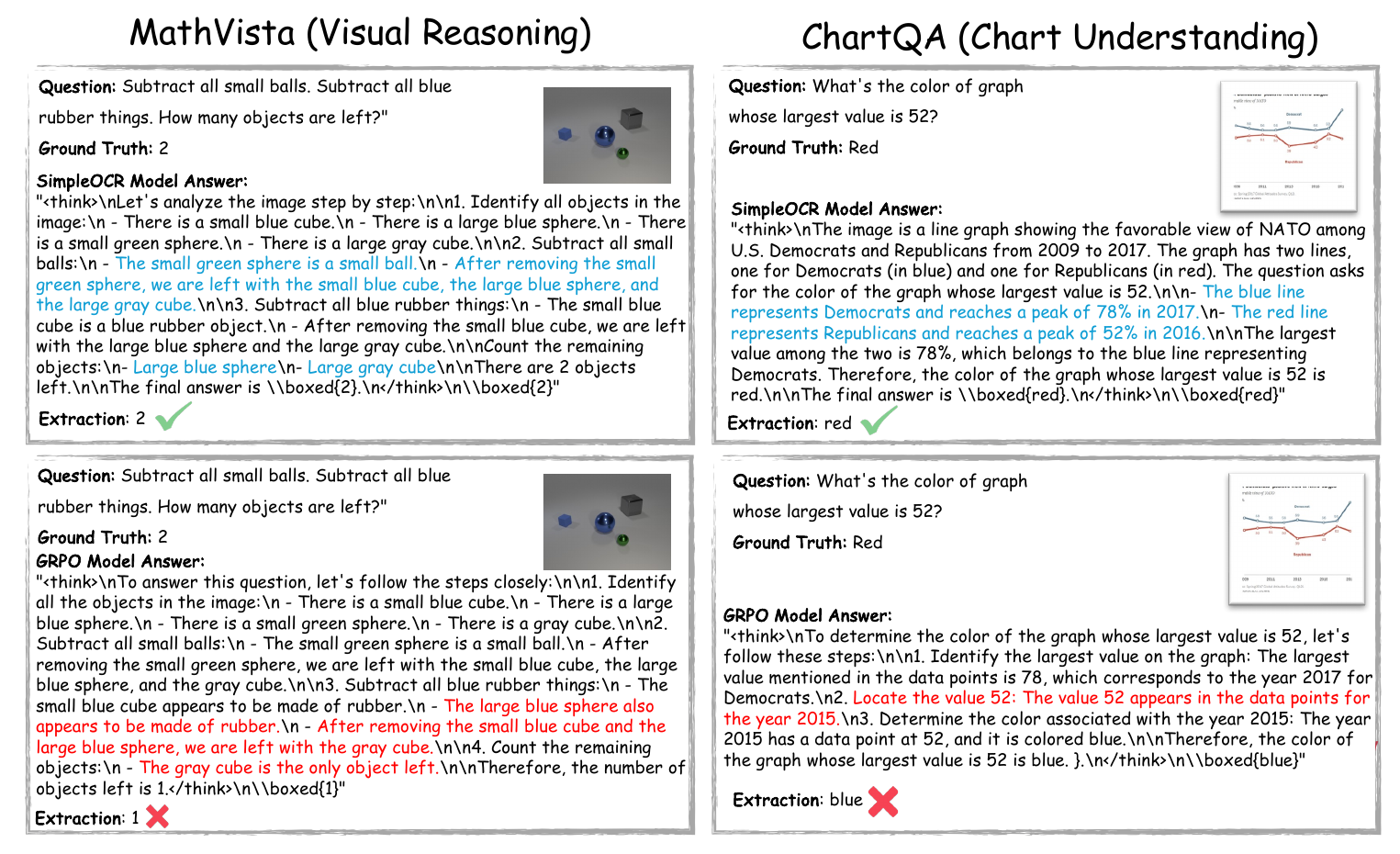}
    \caption{Left: On MathVista, the GRPO baseline is misled by hallucinated semantic priors, while SimpleOCR correctly identifies material properties. Right: On ChartQA, the baseline relies on superficial keyword spotting, whereas SimpleOCR performs holistic visual analysis. \textcolor{blue}{Blue}: correct grounding; \textcolor{red}{red}: heuristic errors.}
    \vspace{-1em}
    \label{fig:case_study}
\end{figure*}

\paragraph{Sensitivity to Group Sampling Size.}
\begin{table}[t]
    \centering
    \caption{Impact of Group Sampling Size $n$. We analyze the effect of the number of generations per prompt during GRPO training.}
    \label{tab:ablation_group_size}
    \resizebox{\columnwidth}{!}{%
    \begin{tabular}{lcccccccccc}
    \toprule
    & \multicolumn{3}{c}{\textbf{In-Domain}} & \multicolumn{6}{c}{\textbf{Out-of-Distribution}} \\
    \cmidrule(lr){2-4} \cmidrule(lr){5-10}
    \textbf{Configuration} & \textbf{Geo3K} & \textbf{MMK12} & \textbf{Avg.} & \textbf{M-Verse} & \textbf{M-Vision} & \textbf{M-Vista} & \textbf{WeMath} & \textbf{Hallusion} & \textbf{Avg.} \\
    \midrule
    $n=3$ & 40.4 & \textbf{62.5} & 51.5 & 46.2 & 24.0 & 67.5 & 60.5 & \textbf{70.4} & 53.7 \\
    $n=6$ & \textbf{43.4} & 62.3 & \textbf{52.9} & \textbf{47.7} & \textbf{24.9} & \textbf{68.7} & \textbf{64.0} & 69.1 & \textbf{54.9} \\
    $n=9$ & 41.4 & 63.0 & 52.2 & 47.4 & 24.6 & 66.4 & 61.6 & 67.9 & 53.6 \\
    \bottomrule
    \end{tabular}%
    }
\end{table}
We investigate the impact of the group size $n$ (the number of rollouts generated per prompt) on SimpleOCR training dynamics in Table~\ref{tab:ablation_group_size}, employing the 7B model as the backbone.
Standard RL scaling laws typically suggest that larger group sizes improve gradient estimation. However, our results reveal an inverted U-shaped trend.
Increasing the group size from $n=3$ to $n=6$ yields a robust 2.2\% gain in average OOD performance, confirming that sufficient exploration is critical for learning complex visual grounding.
Crucially, further scaling to $n=9$ does not yield additional benefits; instead, performance suffers a slight 2.4\% regression.
We hypothesize that in the context of VQ training, excessively large groups may introduce ``reward hacking'' on noisy visual samples or optimization instability. Consequently, we adopt $n=6$ as the optimal trade-off between computational efficiency and reasoning performance.

\subsection{Qualitative Analysis}

Figure~\ref{fig:case_study} illustrates the behavioral shift. In visual reasoning (MathVista), the baseline GRPO model succumbs to semantic priming, associating the text ``blue'' with a prominent sphere despite conflicting visual evidence (metallic luster), whereas SimpleOCR discriminates texture correctly. Similarly, on ChartQA, the baseline relies on superficial keyword spotting, matching ``52'' without comprehending the structural condition ``largest value'', while SimpleOCR successfully parses the chart topology.
These cases validate that the capability-utilization gap is not a deficit of perception but of execution preference. Standard models default to spurious text shortcuts, but SimpleOCR structurally blocks this path, compelling the model to engage in grounded visual reasoning.

\section{Conclusion}
In this paper, we identified and quantified the ``modality laziness'' in MLLMs, where models bypass visual evidence in favor of text-based shortcuts. Our diagnostic VQ setting revealed a significant capability-utilization gap, which we addressed through \ours. By structurally enforcing visual engagement via randomized text rendering, \ours\ effectively transforms the model's reliance from parametric priors to grounded visual perception. Empirically, \ours\ delivers consistent improvements across both in-domain and out-of-distribution benchmarks. Notably, it achieves these gains with extreme data efficiency (using 30$\times$ less data than comparable RL methods) and seamless plug-and-play compatibility with existing frameworks. 

\section*{Acknowledgments}
This work was partially supported by the Amazon Research Award, the Cisco Faculty Research Award.
\section*{Limitations}
While \ours\ effectively bridges the capability-utilization gap, we identify two primary limitations.
First, our method operates as an \textit{elicitation strategy} rather than a fundamental capability builder. It relies on the base MLLM having latent OCR capabilities (i.e., a strong vision encoder) to recognize the rendered text. 
Second, our approach is bounded by \textit{visual resolution constraints} when handling extremely long queries. Unlike text encoders that scale efficiently to long contexts, rendering extensive text prompts (e.g., multi-paragraph instructions) onto a single image is limited by the vision encoder's input resolution.

\noindent\textbf{Potential Risks.} 
Enhanced visual text extraction could theoretically be leveraged to bypass visual security measures (e.g., CAPTCHA solvers) or to automate the extraction of sensitive personal information from natural images (e.g., reading documents or screens in the background of photos). 
However, our method functions as an activation strategy for existing base models rather than introducing new, specialized attack capabilities. The risks are inherently bound by the safety alignment and capabilities of the underlying foundation models.

\bibliography{custom}
\clearpage
\appendix
\label{sec:appendix}

\section{Dataset Details}
\label{app:dataset_details}

\subsection{Training Data}
Our training set consists of two high-quality mathematical reasoning datasets, totaling 8.5K instances. Detailed statistics are provided in Table~\ref{tab:train_data_stats}.

\begin{table}[h]
    \centering
    \caption{\textbf{Training Data Statistics.} We combine geometry-focused and general K-12 math datasets to construct a diverse training corpus.}
    \label{tab:train_data_stats}
    \resizebox{0.9\linewidth}{!}{%
    \begin{tabular}{lccc}
    \toprule
    \textbf{Dataset} & \textbf{Source} & \textbf{Domain} & \textbf{Size} \\
    \midrule
    Geometry3K & \citep{lu2021inter} & Plane Geometry & 2,100 \\
    MMK12 & \citep{meng2025mm} & K-12 Mathematics & 6,400 \\
    \midrule
    \textbf{Total} & - & \textbf{Mixed} & \textbf{8,500} \\
    \bottomrule
    \end{tabular}%
    }
\end{table}

\noindent
\textbf{Geometry3K}~\citep{lu2021inter}. A high-quality geometry problem-solving dataset containing formal geometric diagrams and corresponding problem descriptions. We utilize the training split (2.1K samples) to enhance the model's spatial reasoning and geometric calculation capabilities.

\noindent
\textbf{MMK12}~\citep{meng2025mm}. A comprehensive multimodal dataset derived from K-12 mathematics curriculum. It covers a wide range of topics including algebra, arithmetic, and function analysis. The subset used (6.4K samples) provides diverse visual-text reasoning scenarios essential for general mathematical grounding.

\subsection{Evaluation Benchmarks}
To rigorously assess generalization capabilities, we evaluate on five mathematical reasoning benchmarks and two OCR-intensive tasks.

\paragraph{Mathematical Reasoning.}
\begin{itemize}
    \item \textbf{MathVista}~\citep{lu2023mathvista}. A comprehensive benchmark integrating diverse mathematical reasoning tasks. It serves as a primary gauge for general multimodal mathematical capability.
    \item \textbf{MathVision}~\citep{wang2024measuring}. A large-scale benchmark designed to evaluate MLLMs across diverse mathematical domains and complex visual contexts.
    \item \textbf{MathVerse}~\citep{zhang2024mathverse}. A dataset specifically curated to diagnose whether MLLMs truly interpret visual diagrams or rely on text shortcuts. This aligns perfectly with our study's motivation to detect ``modality laziness''.
    \item \textbf{WeMath}~\citep{qiao2024wemath}. A benchmark focusing on human-like reasoning processes in complex mathematical problems, testing the depth of the model's logical derivation.
    \item \textbf{HallusionBench}~\citep{guan2024hallusionbench}. An advanced diagnostic suite for detecting visual hallucinations and illusions. We use it to verify faithful visual grounding and resistance to perceptual interference.
\end{itemize}

\paragraph{OCR-Intensive Tasks.}
To verify the transfer of visual text reading skills, we include two specific benchmarks:
\begin{itemize}
    \item \textbf{ChartQA}~\citep{masry2022chartqa}. A dataset requiring reasoning over charts with data labels, titles, and legends, serving as a direct test of the model's ability to extract and integrate fine-grained visual text.
    \item \textbf{InfographicVQA}~\citep{mathew2022infographicvqa}. A benchmark challenging models to understand complex document layouts and infographics with high-density text.
\end{itemize}

\section{System Prompts}
\label{app:prompts}

We utilize the standard system prompt from the \texttt{verl} framework to elicit structured reasoning (Chain-of-Thought) and formatted answers.

\begin{tcolorbox}[title={System Prompt for Reasoning}, width=\linewidth, fontupper=\small, colback=gray!5, colframe=black!60]
\small
\textit{``Solve the question. The user asks a question, and you solve it. You first think about the reasoning process in the mind and then provide the user with the answer. The reasoning process MUST BE enclosed within \texttt{<think>} \texttt{</think>} tags. The answer is in latex format and wrapped in \texttt{\$...\$}. The final answer must be wrapped using the \texttt{\textbackslash boxed\{\}} command.''}
\end{tcolorbox}

\subsection{Evaluated Models}
\label{app:evaluated_models}

We include a comprehensive set of state-of-the-art multimodal models in our evaluation, categorized into general-purpose open-source baselines and recent RL-optimized models.

\paragraph{Open-Source Baselines.}
\begin{itemize}
    \item \textbf{Qwen2.5-VL-3/7B-Instruct}~\citep{bai2025qwen2}. 
    The latest iteration of the Qwen-VL series, featuring state-of-the-art OCR and visual understanding capabilities trained on massive-scale datasets. We utilize these as our primary base models to demonstrate the effectiveness of \ours.
    
    \item \textbf{InternVL-2.5-8B-Instruct}~\citep{chen2024expanding}. 
    A powerful MLLM that expands performance boundaries through model and test-time scaling, known for its strong general-purpose visual perception.
    
    \item \textbf{LLaVA-OneVision-7B}~\citep{li2024llava}. 
    A model designed for easy visual task transfer, utilizing a unified architecture to handle diverse vision-language scenarios efficiently.
    
    \item \textbf{Kimi-VL-16B}~\citep{team2025kimi}. A large-scale open-weights model utilizing a Mixture-of-Experts (MoE) architecture, demonstrating competitive performance on chart and document understanding benchmarks.
    
    \item \textbf{Mulberry-7B}~\citep{yao2024mulberry}. 
    An MLLM empowered with OpenAI-o1-like reasoning capabilities via collective Monte Carlo Tree Search (MCTS), focusing on enhanced logical deduction.
\end{itemize}

\paragraph{RL-Optimized \& R1-Series Models.}
\begin{itemize}
    \item \textbf{Math-LLaVA.}~\citep{shi2024math} 
    A specialized model bootstrapped for mathematical reasoning, serving as a strong baseline for SFT-based mathematical capability.
    
    \item \textbf{R1-VL-7B.}~\citep{zhang2025r1} 
    A pioneering model trained via step-wise Group Relative Policy Optimization (GRPO), explicitly rewarding intermediate reasoning steps to improve logical consistency.
    
    \item \textbf{R1-OneVision-7B.}~\citep{yang2025r1} 
    An extension of the R1 series that advances generalized multimodal reasoning through cross-modal formalization techniques.
    
    \item \textbf{ThinkLite-7B-VL.}~\citep{wang2025sota} 
    A data-efficient model achieving state-of-the-art performance with fewer samples, utilizing MCTS-guided sample selection for self-improvement.
    
    \item \textbf{VLAA-Thinker-7B.}~\citep{chen2025sft} 
    A model investigating the trade-offs between SFT and RL in R1-like reasoning, providing insights into training recipes for reasoning-heavy MLLMs.
    
    \item \textbf{MM-Eureka-8B.}~\citep{meng2025mm} 
    A model exploring the frontiers of multimodal reasoning using rule-based reinforcement learning, emphasizing verified feedback signals.
\end{itemize}

\section{Detailed Ablation Results}
\label{app:ablation_results}

In Section 5.4, we discussed the optimization conflict observed in mixed training strategies. Table~\ref{tab:ablation_full_appendix} provides the detailed performance breakdown across four representative out-of-distribution benchmarks. 

As shown, the mixed strategy (50\% VQ) fails to improve over the baseline in most reasoning-intensive tasks (e.g., WeMath, MathVista), confirming that the conflicting modality signals hinder model convergence. In contrast, the pure SimpleOCR strategy (100\% VQ) achieves the best average performance across the board.

\begin{table}[t] 
    \centering
    \caption{Impact of VQ Training Ratio (Detailed Breakdown). We report the performance on four reasoning-heavy OOD benchmarks. The mixed strategy (50\% VQ) consistently underperforms or stagnates compared to the baseline (Avg. 49.3 vs 50.3), supporting the hypothesis of optimization conflict. Only the full VQ strategy (\ours) achieves robust generalization gains (Avg. 51.3).}
    \label{tab:ablation_full_appendix}
    \resizebox{\columnwidth}{!}{
    \begin{tabular}{lcccc|c}
    \toprule
    \textbf{VQ Ratio} & \textbf{MathVerse} & \textbf{MathVision} & \textbf{MathVista} & \textbf{WeMath} & \textbf{Avg.} \\
    \midrule
    Standard (0\% VQ) & 46.4 & 22.5 & 66.9 & 65.3 & 50.3 \\
    Mixed (25\% VQ) & 47.8 & 22.9 & 67.2 & 62.8 & 50.2 \\
    Mixed (50\% VQ) & 46.2 & 23.7 & 65.0 & 62.4 & 49.3 \\
    Mixed (75\% VQ) & 48.0 & 23.9 & 65.7 & 62.2 & 50.0 \\
    \textbf{SimpleOCR (100\% VQ)} & \textbf{47.7} & \textbf{24.9} & \textbf{68.7} & 64.0 & \textbf{51.3} \\
    \bottomrule
    \end{tabular}%
    }
\end{table}

\section{OCR-Intensive Benchmarks}
\label{app:ocr_results}

We provide the exact numerical breakdown for OCR-intensive tasks in Table~\ref{tab:ocr_results}. 
A key observation is that standard GRPO can lead to negative transfer on fine-grained visual tasks like ChartQA (dropping from 79.8\% to 79.5\%), likely due to the model overfitting to textual reasoning shortcuts. 
In contrast, \ours\ consistently yields improvements across all metrics (81.6\% on ChartQA), confirming its effectiveness in preserving and enhancing visual grounding capabilities without compromising general reasoning.

\begin{table}[h]
    \centering
    \caption{\textbf{Performance on OCR-Intensive Benchmarks.} Exact numbers corresponding to Figure~\ref{fig:ocr_benchmark}.}
    \label{tab:ocr_results}
    \resizebox{\columnwidth}{!}{%
    \begin{tabular}{lccc}
    \toprule
    \textbf{Method} & \textbf{ChartQA} & \textbf{HallusionBench} & \textbf{InfoVQA} \\
    \midrule
    Base Model & 79.8 & 68.2 & 79.7 \\
    GRPO (Original images) & 79.5 & 68.9 & 80.1 \\
    GRPO + SimpleOCR & \textbf{81.6} & \textbf{69.1} & \textbf{80.5} \\
    \bottomrule
    \end{tabular}%
    }
\end{table}

\section{Evaluation Protocol Details}
\label{app:eval_protocol}

\paragraph{Inference and Extraction.}
For all experiments, we perform inference using greedy decoding (\texttt{temperature=0}) to ensure reproducibility. 
To isolate the final answer from the Chain-of-Thought (CoT) rationale, we employ a rule-based extraction parser. Specifically, we extract the content within the last occurrence of the \texttt{\textbackslash boxed\{\dots\}} delimiter in the model output. If no such delimiter is found, the raw output is passed to the subsequent evaluation stages.

\paragraph{Hierarchical Judging Pipeline.}
We implement a two-stage cascaded evaluation strategy to balance strict symbolic correctness with semantic flexibility:

\begin{enumerate}
    \item \textbf{Stage 1: Symbolic Verification (Math-Verify).} 
    We first employ the \texttt{math-verify} library for symbolic equivalence checks. This tool parses mathematical expressions into canonical forms (e.g., standardizing fractions, square roots, and units) to determine correctness. If \texttt{math-verify} returns a positive match, the sample is marked as correct immediately.

    \item \textbf{Stage 2: LLM-based Fallback Judge.} 
    For samples where symbolic verification fails or is inconclusive (e.g., complex textual reasoning or format mismatches), we employ \texttt{gpt-4o-2024-08-06} as a fallback evaluator. We construct a meta-evaluation prompt containing the question, the ground truth, and the student's answer. The LLM is strictly instructed to:
    \begin{itemize}
        \item Ignore superficial formatting differences (e.g., Markdown styling).
        \item Check for mathematical equivalence rather than string matching.
        \item Allow a relative numerical tolerance of $\pm 1\%$ (unless specified otherwise).
        \item For multiple-choice questions, verify that the selected option letter matches the ground truth.
    \end{itemize}
    The LLM outputs a binary score (0 or 1) based on these criteria.
\end{enumerate}

\paragraph{Benchmark-Specific Protocols.}
\begin{itemize}
    \item \textbf{HallusionBench:} We strictly adhere to the official evaluation protocol, utilizing its dataset-specific LLM judge to handle the unique ``uncertain'' label requirements.
    \item \textbf{Geometry3K:} Due to the strict formatting of this dataset, we rely primarily on symbolic verification, enforcing exact matches for geometric values and units.
\end{itemize}

\section{Supplementary Implementation Details}
\label{app:implementation_details}

We provide the detailed hyperparameter configurations used in our experiments in Table~\ref{tab:hyperparams}.

\begin{table}[h]
    \centering
    \caption{\textbf{Summary of hyperparameter configurations.}}
    \label{tab:hyperparams} 
    \resizebox{\columnwidth}{!}{
    \begin{tabular}{lc}
    \toprule
    \textbf{Parameter} & \textbf{Configuration} \\
    \midrule
    Model Base & Qwen2.5-VL-Instruct \\
    Vision Encoder & Frozen \\
    Global Batch Size & 128 \\
    Rollout Batch Size & 512 \\
    Rollout Temperature & 1.0 \\
    Learning Rate & $1\times 10^{-6}$ \\ 
    Optimizer & AdamW \\
    Total Training Steps & 200 \\
    CPU Memory & 512GB \\
    GPU & RTX 6000 Pro Blackwell \\ 
    \bottomrule
    \end{tabular}%
    }
\end{table}

\end{document}